\def\year{2019}\relax
\documentclass[letterpaper]{article}

\usepackage{aaai}
\usepackage{times}
\usepackage{helvet}
\usepackage{courier}
\usepackage{amssymb}
\usepackage{amsmath}
\usepackage{amsthm}
\usepackage{color}
\usepackage[utf8]{inputenc}
\usepackage{graphicx}
\usepackage{dsfont}
\usepackage{algorithmicx}
\usepackage{algpseudocode}
\usepackage{xspace}
\usepackage{natbib}
\usepackage{hhline}
\usepackage[hyphens]{url}
\usepackage{comment}
\usepackage[makeroom]{cancel}
\DeclareMathOperator*{\argmax}{argmax}

\usepackage{subcaption}

\usepackage[compact]{titlesec}         
\titlespacing{\section}{0pt}{5pt}{0pt} 

\title{Transfer Learning Across Simulated Robots With Different Sensors}

\author{
	H\'el\`ene Plisnier,\textsuperscript{1}
	Denis Steckelmacher,\textsuperscript{1}
	Diederik M. Roijers,\textsuperscript{2}
	Ann Now\'e\textsuperscript{1} \\
	\textsuperscript{1} Vrije Universiteit Brussel \\
	\textsuperscript{2} Vrije Universiteit Amsterdam
}

\allowdisplaybreaks
\newcommand{\pis}{\ensuremath{\pi_{\text{source}}}\xspace}

\begin{document}

\maketitle

\begin{abstract}

For a robot to learn a good policy, it often requires expensive equipment (such as sophisticated sensors) and a prepared training environment conducive to learning. However, it is seldom possible to perfectly equip robots for economic reasons, nor to guarantee ideal learning conditions, when deployed in real-life environments. A solution would be to prepare the robot in the lab environment, when all necessary material is available to learn a good policy. After training in the lab, the robot should be able to get by \emph{without} the expensive equipment that used to be available to it, and yet still be guaranteed to perform well on the field.
The transition between the lab (\emph{source}) and the real-world environment (\emph{target}) is related to transfer learning, where the state-space between the source and target tasks differ. We tackle a simulated task with continuous states and discrete actions presenting this challenge, using Bootstrapped Dual Policy Iteration, a model-free actor-critic reinforcement learning algorithm, and Policy Shaping. 
Specifically, we train a BDPI agent, embodied by a virtual robot performing a task in the V-Rep simulator, sensing its environment through several proximity sensors. The resulting policy is then used by a second agent learning the same task in the same environment, but with camera images as input. The goal is to obtain a policy able to perform the task relying on merely camera images.


\end{abstract}

\section{Introduction}
A more and more viable alternative to hand-coding the whole behavior of a robot, is to let the robot learn by itself how it should act in order to reach a human-defined goal, using Reinforcement Learning (RL) \citep{Sutton2018}. However, autonomous robots generally require significant equipment allowing them to sense their surroundings, such as various proximity sensors, and might need their whole environment to be designed so to ease their navigation, and favour smooth operation. Unfortunately, it is often unfeasible: i) to equip robots targeted to a general audience with expensive sensory equipment, and ii) to ensure the best possible conditions in any real-life situations, e.g., preparing a user's house, or the streets the robot needs to navigate in. It is however possible to provide a prototypical robot with the necessary sensors, as well as a fitting environment within the robot designers' lab. Nevertheless, to guarantee good performance outside the lab, it is crucial that the robot learns to get by without the help of sensors and exceptionally good environmental conditions. This particular problem of transitioning from the lab world to the real world falls into the realm of \emph{transfer learning}.

Transfer learning has the potential to make RL agents a lot faster at mastering new tasks, by allowing the agent to reuse knowledge acquired in one or several previous tasks. Many different components can vary between the source and target tasks. The state description, for instance, might not be the same in the two tasks; one might be richer and/or easier to learn from, than the other \citep[Section 3.2.1]{Taylor2009}. This dissimilarity naturally emerges when trying to share knowledge across robots equipped with different sensors. In this paper, we are interested in transferring knowledge from a robotic platform to another, both tackling the same task with the same action set, while sensing their environment differently. The transfer is made \emph{from} the robot which state description is empirically easier to learn from (output of 8 proximity sensors) \emph{to} the robot which state description is harder to learn from (output of a single camera). This could allow a cheap under-equipped robot to perform as well on the field as a more sophisticated one, for which the task is much easier to learn.

Current techniques to transfer learning can be sorted in two categories: the ones that use the transferred knowledge to bias the agent's exploration strategy \citep{Fernandez2006, Taylor2007a, Madden2004, Zhan2015, Plisnier2019}, a technique primarily used with Q-Learning-based methods; the ones that train the agent to imitate the transferred knowledge, either by initializing the agent with the transferred knowledge \citep{Taylor2007b} or by dynamically teaching the agent \citep{Parisotto2015, Brys2015, Plisnier2019}. Except for the Actor-Advisor \citep{Plisnier2019}, there has not been much investigation on allowing the agent to both be guided by transferred knowledge, and learn to imitate it at the same time. In addition, to our knowledge, transfer between tasks with completely different state-spaces consists in a novel setting.


Our agent, an Epuck robot simulated in V-REP \footnote{V-REP is a software allowing the simulation of several existing robots in an environment that can be customized by algorithms designers: http://www.coppeliarobotics.com/index.html.}, learns using Bootstrapped Dual Policy Iteration \citep[BDPI]{Steckelmacher2019}. BDPI is a model-free actor-critic reinforcement learning algorithm for continuous states and discrete actions settings, with one actor and several off-policy critics. This algorithm allows us to investigate three forms of transfer, inspired by the Actor-Advisor: i) purely via exploration alteration, ii) purely via imitation learning, and iii) a mix of both exploration alteration and imitation learning. We empirically show that our BDPI extension for transfer learning allows a simulated camera-equipped Epuck robot to leverage proximity sensors that are not at its disposal, and learn a much better policy than what was originally possible to achieve with its simple setup. 


\section{Background}
\label{sec:background}

\subsection{Markov Decision Processes}
\label{sec:background_mdp}

A discrete-time Markov Decision Process (MDP) \citep{Bellman1957} with discrete actions is defined by the tuple $\langle S, A, R, T \rangle$: a possibly-infinite set $S$ of states; a finite set $A$ of actions; a reward function $R(s_t, a_t, s_{t+1}) \in  \mathbb{R}$ returning a scalar reward $r_t$ for each state transition; and a transition function $T(s_{t+1} | s_t, a_t) \in [0, 1]$ taking as input a state-action pair $(s_t, a_t)$ and returning a probability distribution over new states $s_{t+1}$.

A stochastic stationary policy $\pi(a_t | s_t) \in [0, 1]$ maps each state to a probability distribution over actions. At each time-step, the agent observes $s_t$, selects $a_t \sim \pi(s_t)$, then observes $r_{t+1}$ and $s_{t+1}$. The $(s_t, a_t, r_{t+1}, s_{t+1})$ tuple is called an \emph{experience} tuple. An optimal policy $\pi^*$ maximizes the expected cumulative discounted reward $E_{\pi^*}[\sum_t \gamma^t r_t]$, where $\gamma$ is a discount factor. The goal of the agent is to find $\pi^*$ based on its experiences within the environment.

\subsection{Bootstrapped Dual Policy Iteration}
\label{subsec:bdpi}

Bootstrapped Dual Policy Iteration \citep[BDPI]{Steckelmacher2019} is an actor-critic method, with one actor and $N_c > 1$ critics. The critics are trained using Aggressive Bootstrapped Clipped DQN \citep{Steckelmacher2019}, a version of Clipped DQN \citep{Fujimoto18} that performs $N_t > 1$ training iterations per training epoch. Each critic maintains two Q-functions, $Q^A$ and $Q^B$. Each training epoch, a batch $b_i$ is sampled for each critic $i \in [1, N_c]$ from an experience buffer $B$. Then, for each training iteration, every critic $i$ swaps its $Q^A$ and $Q^B$ functions, then $Q^A$ is trained using Equation \ref{eq:abcdqn} on $b_i$.
	
\begin{align} 
	\label{eq:abcdqn}
	Q_{k+1}(s_t, a_t) &= Q_k(s_t, a_t) \,  + \, \alpha \, (r_{t+1} + \gamma \, V(s_{t+1})) \\ &- Q_k(s_t, a_t) \\
	V(s_{t+1})) &= \min_{l = A, B} Q^l (s_{t+1}, \argmax_{a'} Q^A(s_{t+1} ,a')) 
\end{align}

The actor $\pi$ is trained using a variant of Conservative Policy Iteration \citep{Pirotta2013}. Every training epoch, after the critics have been updated for a number $N_t$ of times, the actor is trained towards the greedy policy of all its critics. This is achieved by sequentially applying Equation \ref{eq:actor} $N_c$ times, each iteration updating the actor based on a different critic.

\begin{align} 
	\label{eq:actor}
	\pi(s) &\leftarrow (1-\lambda) \, \pi(s) + \lambda \, \Gamma(Q^{A, i}_{k+1}(s, \cdot))     &    \forall s \in b_i
\end{align}

\noindent where $\lambda = 0.05$. A great asset of BDPI over other state-of-the-art actor-critic methods is its high sample-efficiency, due to the aggressiveness of its off-policy critics.

\subsection{Policy Shaping and the Actor-Advisor}
\label{sec:background_ps_aa}

Policy Shaping \citep{Kartoun2010, Griffith2013, Macglashan2017, Harrison2018} generally aims at letting an external advisory policy \pis (we call it \pis since, in our case, it is learned in the \emph{source} task) alter or determine the agent's behavior. The specific Policy Shaping formula we are considering in this paper is the one suggested by \cite{Griffith2013}:

\begin{align} 
    \label{eq:ps}
    a_t \sim & \underbrace{
        \frac{\pi(s_t) \pis(s_t)} {\pi(s_t) \cdot \pis(s_t)}
    }_{
        \sum_{a \in A} \pi(a | s_t) \pis(a | s_t)
    }
\end{align}

\noindent where $\pi(s_t)$ is the state-dependent policy learned by the agent, $\pis(s_t)$ is the state-dependent advice, and $\pi(s_t) \cdot \pis(s_t)$ is the dot product. The actions executed by the agent in the environment are sampled from a mixture of the agent's current learned policy $\pi$ and an external advisory policy $\pis$. Executing actions from this mixture allows the advisor $\pis$ to guide the agent's exploration and potentially improves its performance. This method not only allows the actor to benefit from the advisor's expertise; it also lets the actor eventually outperform its advisor, if the actor's sensors are more informative than the advisor's ones. This way, the actor's performance is never bounded by its advisor's, and the advisor does not need to have a complete knowledge of the task to be solved. 

The Actor-Advisor \citep{Plisnier2019} applies this technique to a variety of RL sub-domains, namely learning from a human teacher, learning under a safe backup policy, and transferring a previously learned policy. The Actor-Advisor assumes a Policy Gradient \citep{Sutton2000} actor, learning a parametric policy $\pi_{\theta}$. At acting time, actions are sampled from the mixed policy $\pi_{\theta}(s_t, \pis(s_t))$, obtained with the above-mentioned policy shaping formula in Equation \ref{eq:ps} \citep{Griffith2013}. The mixed policy $\pi_{\theta}(s_t, \pis(s_t))$ is integrated in the standard Policy Gradient loss \citep{Sutton2000} used to train the agent:

\begin{align}
    \label{eq:pgg}
    \mathcal{L}(\pi) &= -\sum\limits_{t=0}^{T} \mathcal{R}_t \log (\pi_{\theta}(a_t | s_t, \pis(s_t)))
\end{align}

\noindent with $\pi_{\theta}(a_t | s_t, \pis(s_t))$ the probability to execute action $a_t$ at time $t$, given as input the state $s_t$ and some state-dependent advice $\pis(s_t)$, and the return $\mathcal{R}_t = \sum_{\tau=t}^{T} \gamma^{\tau} r_{\tau}$, with $r_\tau = \mathcal{R}(s_\tau, a_\tau, s_{\tau+1})$, a simple discounted sum of future rewards. Computing the gradient based on the mixed policy $\pi_{\theta}(s_t, \pis(s_t))$ rather than on $\pi_{\theta}$ only potentially allows the actor's learning to be influenced by the advisor, in addition to being guided during exploration.

\section{Transfer Learning}
\label{sec:sota}

Transferring knowledge in Reinforcement Learning potentially improves sample-efficiency, as it allows an agent to exploit relevant past knowledge while tackling a new task, instead of learning it from scratch \citep{Taylor2009}. Usually, we consider that the valuable knowledge to be transferred in Reinforcement Learning is the actual output of a reinforcement learner: an action-value function $Q(s, a)$ or a policy $\pi$ \cite[p. 34]{Brys2016}. Some work also consider reusing skills, or \emph{options} \citep{Sutton1999}, as a transfer of knowledge across tasks \citep{Andre2002, Ravindran2003, Konidaris2007}. In this section, we sort previous work in categories related to the \emph{way} $\pis$ is transferred into the agent, and look at what is allowed to be different between the source task and the target task. The two predominant ways in which knowledge can be transferred are i) $\pis$ serves as a guide during exploration, ii) $\pis$ is used to train or initialize the agent, so that the agent actively imitates $\pis$.


\subsubsection{Exploration}

A fast and effective way of transferring a policy is to leverage it in the agent's exploration strategy. Altering the exploration strategy is a popular technique in the safe RL domain, and consists in biasing or determining the actions taken by the agent at action selection time \citep{Garcia2015}. Such exploration requires the agent to be able to learn from \emph{off-policy} experiences. The motivation behind guided exploration is the poor performance of a fresh agent at the beginning of learning, in addition to the presence of obstacles difficult to overcome in the environment. An exploration guided by a smarter external policy, such as $\pis$, could help improve the agent's early performance, as well as help it in the long run explore fruitful areas. 

Some existing work applies guided exploration to transfer learning \citep{Fernandez2006, Taylor2007b, Madden2004}, and illustrates how this technique allows the agent to outperform $\pis$'s performance. Regarding the components of the source and target tasks that are allowed to differ, \citet{Fernandez2006} considers different goal placements (hence, different reward functions), \citet{Madden2004} uses symbolically learned knowledge to tackle states that are seen by the agent for the first time, and \citet{Taylor2007b} assumes similar state variables and actions, but a different reward function. The translation functions required to map a state/action in one task to a state/action in another are assumed to be provided.



\subsubsection{Learning}

Although an improved exploration might result in an improved performance, and a jump-start occurs, an agent which actions are simply overridden by an external policy does not actively learn to imitate it. Other techniques have proposed to ``teach" the agent to perform the target task (instead of merely guiding it), either by dynamic teaching, or by straightforward initialization. Imitation learning aims to allow a student agent to learn the policy of a demonstrator, out of data that it has generated \citep{Hussein2017}. Similarly, policy distillation \citep{Bucila2006} can be applied to RL to train a fresh agent with one or several expert policies, hence resulting in one, smaller, potentially multi-task RL agent \citep{Rusu2015}.

Imitation learning and policy distillation are somewhat related to transfer learning \citep[p.24]{Hussein2017}, although imitation and distillation assume that the source and target tasks are the same, while transfer does not. The Actor-Mimic \citep{Parisotto2015} uses several DQN policies (each expert in a different source task) to train a multi-task student network, by minimizing the cross-entropy loss between the student and experts' policies. To perform transfer, the resulting multi-task expert policy is used to initialize yet another DQN network, which learns the target task. The Actor-Mimic assumes that the source and target tasks share the same observation and action spaces, with different reward and transition functions. In \emph{Q-value reuse} \citep[Section 5.5]{Taylor2007a}, a Q-Learner uses $Q_{source}$ to kickstarts its learning of the target task, while also learning a new action-value function $Q_{target}$ to compensate $Q_{source}$'s irrelevant knowledge. In \citet{Taylor2007a}, the agent learns to play Keepaway games, and is introduced to a game with more players, resulting in more actions and state variables. \citet{Brys2015} transfers $\pis$ to a Q-Learning agent through reward shaping; the differences in the action and state spaces between source and target tasks are solved using a provided translation function.

The Actor-Advisor \citep{Plisnier2019} tries to get the best of both the exploration alteration and the teaching worlds. It mixes $\pis$ with a Policy Gradient actor's policy at action selection time (hence biasing the exploration strategy), using the policy mixing formula in \citet{Griffith2013}; the mixed policy is also integrated in the actor's loss. This way, the learning process is influenced by $\pis$, while it also guides the agent's exploration. In the transfer task in \citet{Plisnier2019}, the doors of a maze are shifted, resulting in a change in the dynamics of the environment.


\section{BDPI with Transfer Learning}
\label{sec:contribution}
We explore the three transfer learning approaches that can take place while using BDPI as learning algorithm. The first method consists in the transfer of $\pis$ purely via the agent's exploration strategy at action selection time, for which we reuse the policy shaping formula by \cite{Griffith2013} (Section \ref{sec:background_ps_aa}, Equation \ref{eq:ps}). Therefore, at acting time, the agent executes an action sampled from the mixture of the BDPI agent's policy $\pi(s_t)$ with $\pis$. In the rest of this paper, we refer to this transfer learning approach as ``transfer at acting time".


The second method to induce transfer learning is by allowing the agent to actively learn to imitate $\pis$. This is achieved by including the policy mixing in the BDPI actor's update rule (Section \ref{subsec:bdpi}); we refer to this approach as ``transfer at learning time":

\begin{align}
    \label{eq:learn}
    \nonumber
    \pi(s) &\leftarrow (1-\lambda) \, \pi(s) + \lambda \, (D \, \frac{\pi(s) \, \pis(s)} {\pi(s_t) \cdot \pis(s_t)} 
    \nonumber
    \\ &+ E \, \Gamma(Q^{A, i}_{k+1}(s, \cdot)) )     &    \forall s \in b_i
    \nonumber \\
    \nonumber
    & \Leftrightarrow \\
    \nonumber
    \pi(s) &\leftarrow A \, \pi(s) + B \, \frac{\pi(s) \, \pis(s)} {\pi(s_t) \cdot \pis(s_t)}  \\ &+ C \, \Gamma(Q^{A, i}_{k+1}(s, \cdot))     &    \forall s \in b_i
\end{align}

\noindent $C$ is always set to 0.05, $B = (1 - C) \times tl$, where $tl$ denotes a fixed transfer learning parameter, and $A = 1 - B - C$. When $tl = 0$, no transfer at learning time occurs, and it amounts to applying the original BDPI actor update (Equation \ref{eq:actor}). When evaluating transfer at learning time in our experiments, we generally fix $tl$ so that $B < C$, allowing the critics' greedy policy to have a more important influence on the actor. Still, even with a small $B$, our empirical results show that the transfer of $\pis$ has a non-negligible impact on the agent's learning.

\section{Experiments}
\label{sec:experiments}
In our setting, an Epuck robots first learns $\pis$, the optimal policy to navigate in a room while observing the output of several proximity sensors. $\pis$ is then transferred to another Epuck robot, learning the same task, while observing its environment through a single camera. Based on the transfer learning techniques allowed by BDPI (Section \ref{sec:contribution}), we compare three approaches to transfer learning: i) performing the transfer purely at acting time, ii) performing the transfer purely at learning time, and iii) performing the transfer both at acting and learning time. We now detail the environment in which our experiments take place.

\subsubsection{Environment}
\label{subsec:env}
\begin{figure}[t]
	\centering
	\includegraphics[width=85mm]{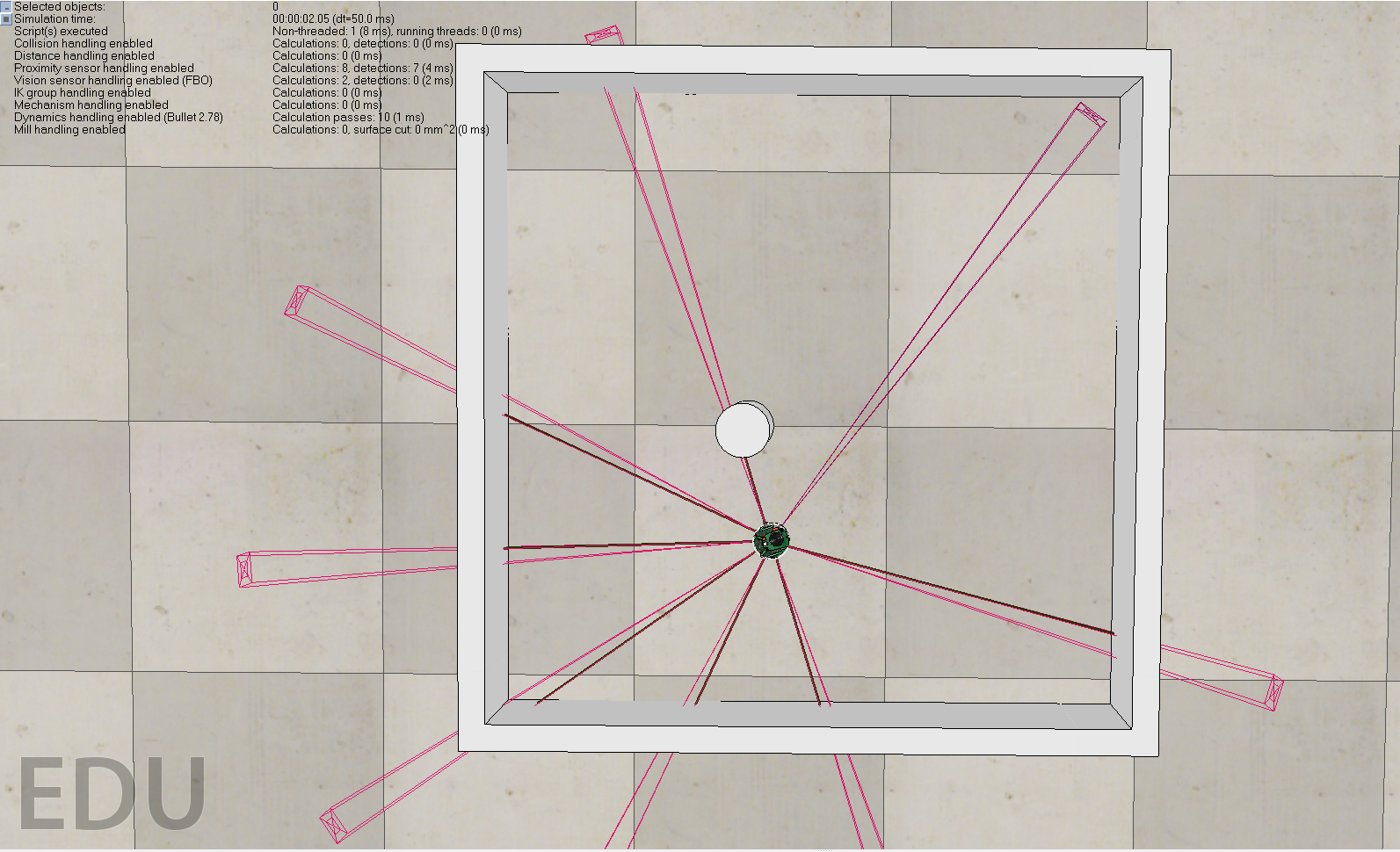}
	\includegraphics[width=85mm]{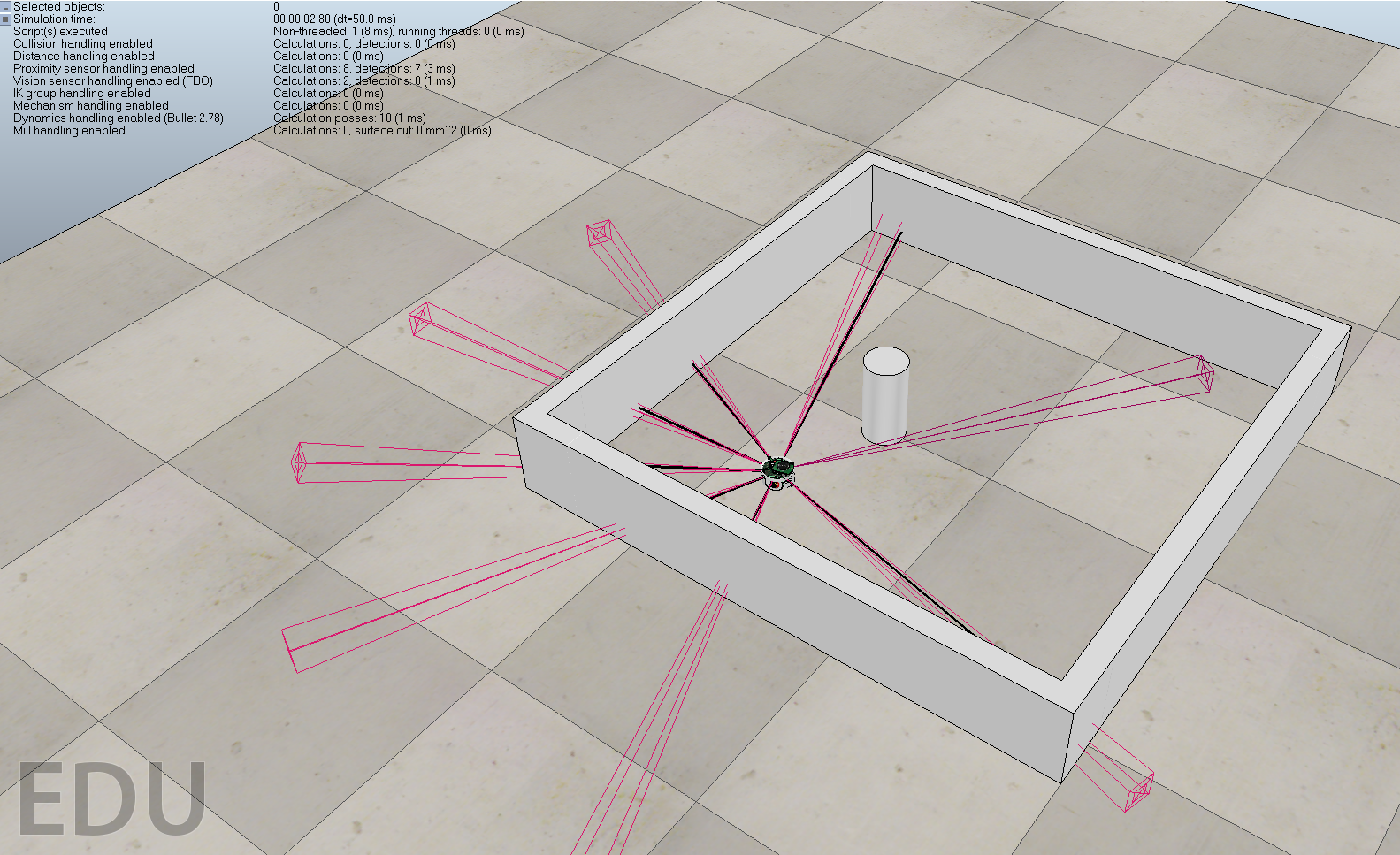}
	\caption{Our simulated Epuck robot learning to navigate in a squared room twice: once out of the output of 8 proximity sensors, and once out of raw camera images, but helped by the policy learned with sensors.}
	\label{fig:env}
\end{figure}

Our agent is embodied by a simulated Epuck robot, in V-REP (Figure \ref{fig:env}). It has 2 actuators, one for its left motor and one for its right motor; 8 proximity sensors, spread around the robot, and one camera on the front. Two environments emerge from this setting: one in which the robot observes the scene through its 8 proximity sensors, and one in which it observes the scene through its camera. In both environments, states are continuous. There are 5 discrete actions: staying still, accelerating the left motor, accelerating the right one, accelerating both, or decelerating both. The robot is in a squared room made of 4 one meter-long walls, and an uncentered pillar. An episode starts with the robot appearing in a random position in the room. Even when the agent only observes camera images, the proximity sensors are used to evaluate how close the robot is to the wall; if its distance to the wall is smaller than 10 cm, it receives a -1 reward. The only positively rewarded action is accelerating both motors (i.e., going forward); if the agent chooses this action, it receives +1. An episode ends after 500 timesteps. The optimal policy is to move in circles around the room without hitting the walls nor the pillar. 

Learning the task merely out of camera images, on the other hand, is much more challenging (impossible, even) for state-of-the-art RL algorithms than from 8 proximity sensors. Hence, the idea is to first learn $\pis$, a near-optimal policy that BDPI learns out the 8 proximity sensors, in less than blah episode, and then to transfer $\pis$ to the agent learning out of camera images. For the transfer of $\pis$ to occur, either at acting or learning time, $\pis$ must still be fed with the output of the 8 proximity sensors. This is achieved by allowing the function \texttt{step()} of our environment to return this output via the ``info" dictionary; hence, our environment returns both the camera images-based observation, and the proximity sensors-based observation to the agent. Therefore, no translation function \citep{Taylor2007a} between state representations is required. Moreover, this implementation remains aligned with our case-scenario, in which robots learn in a lab environment with all necessary sensors input constantly available to them. 

\subsubsection{Setup}
We parametrize BDPI with $N_c = 16$ critics, all trained at every timestep on a new 256 experiences replay minibatch, by applying once ($N_t = 1$) Equation \ref{eq:abcdqn}. The experience minibatches are sampled from a shared 50000 experiences buffer. The critics' learning rate $\alpha$ in Equation \ref{eq:abcdqn} is set to 0.2. 

To first learn $\pis$ based on the proximity sensors (without any transfer), the actor learning rate $\lambda$ in Equation \ref{eq:actor} is set to 0.05, and $tl = 0$. BDPI' neural network is trained for 20 epochs per training iteration, on the mean-squared-error loss. When learning while observing camera images, the number of training epochs is reduced to 1. The policy network has one hidden layer of 100 neurons, with a learning rate of $1e-4$ when learning out of the 8 proximity sensors, and of $1e-6$ when learning out of camera images. 

In the next section, we evaluate three ways to transfer $\pis$ to the BDPI agent while it is learning using only its camera. When transferring $\pis$ solely at acting time, $tl$ is set to 0; when transferring $\pis$ solely at learning time, we set $tl$ to $0.03$; when transfer occurs both at acting and learning time, $tl=0.01$.  

\subsection{Results}
We performed three runs for each of the 5 following settings: i) learning out of proximity sensors, no transfer; ii) learning out of camera images, no transfer; iii) learning out of camera images, with transfer solely at acting time; iiii) learning out of camera images, with transfer solely at learning time; iiiii) learning out of camera images, with transfer both at acting and learning time.

\begin{figure}[t]
	\centering
	\includegraphics[width=85mm]{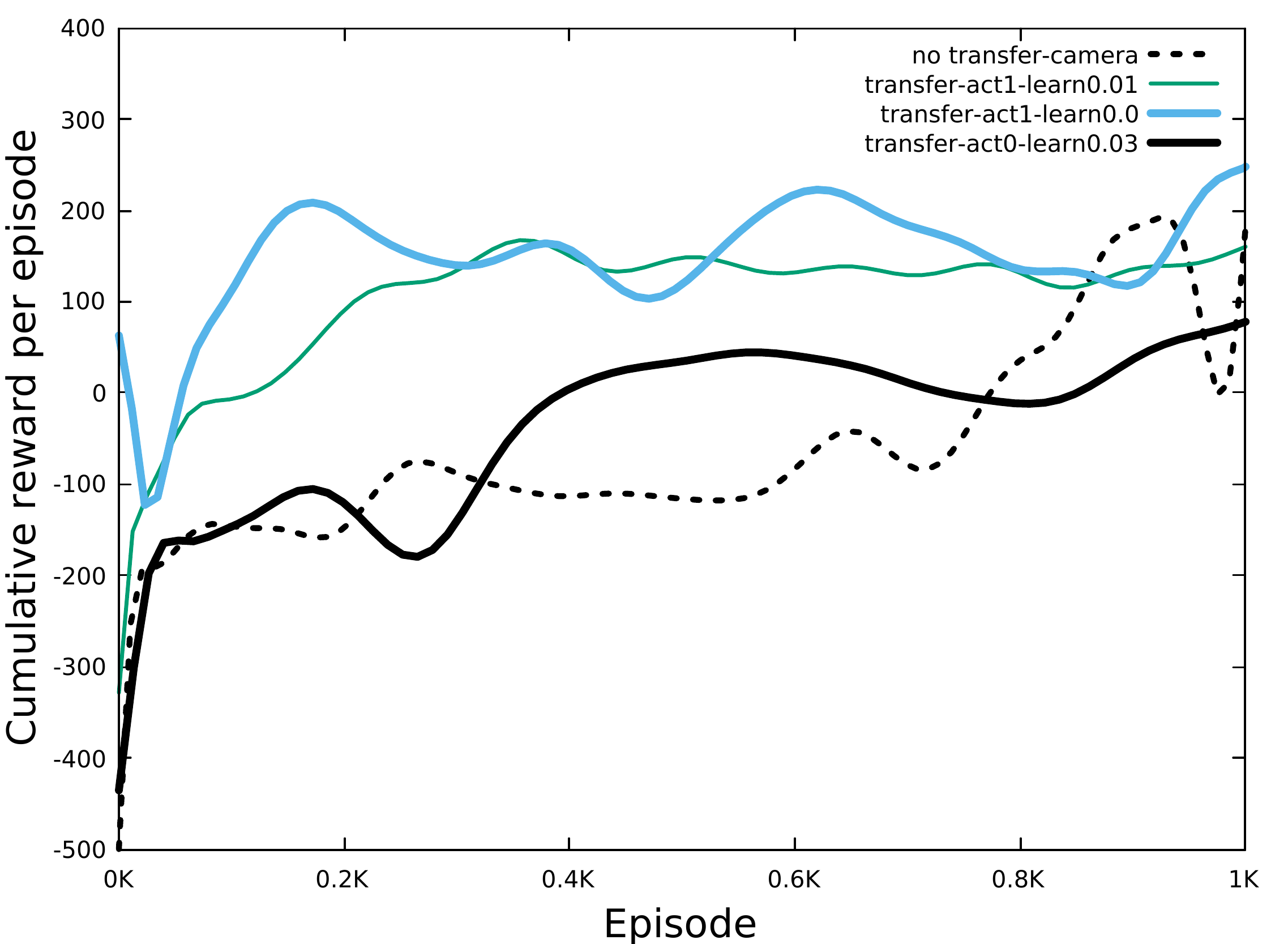}
	\caption{Transfer learning applied: both at acting and learning time (``act1-learn0.01"); solely at acting time (``act1-learn0.0"); solely at learning time (``act0-learn0.03"), compared to no transfer at all while learning with the camera (``no transfer-camera"). Our results, averaged over 3 runs, suggest that transfer only at acting time is the most effective form of transfer, closely followed by the mixed approach.}
	\label{fig:results}
\end{figure}

Transferring $\pis$ solely at acting time, hence strongly altering the agent's exploration strategy, leads to a significant improvement of the agent's performance while learning from camera images (Figure \ref{fig:results}). Compared to the other two approaches, it is also the most effective one. The lesser performance of the mixed approach, compared to the transfer at acting time only one, suggests that allowing $\pis$ to also influence the agent's learning rule can actually be detrimental.

\section{Conclusion}
When deploying physical reinforcement learning agent on the field, it is not always possible to ensure optimal equipment and learning conditions to the agent, as it can be the case in a lab environment. Hence, it is desirable to somehow prepare the agent while in the lab, where all necessary equipment is still available. This way, the agent could learn to \emph{get by without} the particular equipment that will no longer be present once deployed in the real-world. In this paper, we propose to use transfer learning to perform this transition from the lab world to the real-world. A robot which quickly learns the task, thanks to its many expensive sensors that are easy to learn from, helps a lesser equipped robot that senses its environment through raw images. Once the camera-equipped robot has learned the task while being advised by the sensors-equipped one, it performs much better than it would have without transfer, as if it was exploiting expensive sensors that it does not have, and is ready for deployment. To achieve this transfer of policy, we extended BDPI, which allows for three different forms of transfer. Our experiments, simulated in V-REP, show that our method greatly improves the sample-efficiency of an Epuck robot sensing its environment through a single camera, which still consists in a highly challenging problem for state-of-the-art reinforcement learning algorithms.

\bibliographystyle{apalike}
\bibliography{biblio}

\begin{thebibliography}{}

\bibitem[Andre and Russell, 2002]{Andre2002}
Andre, D. and Russell, S.~J. (2002).
\newblock State abstraction for programmable reinforcement learning agents.
\newblock In {\em AAAI/IAAI}, pages 119--125.

\bibitem[Bellman, 1957]{Bellman1957}
Bellman, R. (1957).
\newblock {A Markovian decision process}.
\newblock {\em {Journal Of Mathematics And Mechanics}}.

\bibitem[Brys, 2016]{Brys2016}
Brys, T. (2016).
\newblock {\em Reinforcement Learning with Heuristic Information}.
\newblock PhD thesis, PhD thesis, Vrije Universitet Brussel.

\bibitem[Brys et~al., 2015]{Brys2015}
Brys, T., Harutyunyan, A., Taylor, M.~E., and Now{\'e}, A. (2015).
\newblock Policy transfer using reward shaping.
\newblock In {\em Proceedings of the 2015 International Conference on
  Autonomous Agents and Multiagent Systems}, pages 181--188. International
  Foundation for Autonomous Agents and Multiagent Systems.

\bibitem[Bucila et~al., 2006]{Bucila2006}
Bucila, C., Caruana, R., and Niculescu-Mizil, A. (2006).
\newblock Model compression: Making big, slow models practical.
\newblock In {\em {International Conference on Knowledge Discovery and Data
  Mining}}.

\bibitem[Fern{\'a}ndez and Veloso, 2006]{Fernandez2006}
Fern{\'a}ndez, F. and Veloso, M.~M. (2006).
\newblock Probabilistic policy reuse in a reinforcement learning agent.
\newblock In {\em {International Conference on Autonomous Agents and Multiagent
  Systems}}.

\bibitem[Fujimoto et~al., 2018]{Fujimoto18}
Fujimoto, S., van Hoof, H., and Meger, D. (2018).
\newblock Addressing function approximation error in actor-critic methods.
\newblock In {\em International Conference on Machine Learning}.

\bibitem[Garc{\'i}a and Fern{\'a}ndez, 2015]{Garcia2015}
Garc{\'i}a, J. and Fern{\'a}ndez, F. (2015).
\newblock A comprehensive survey on safe reinforcement learning.
\newblock {\em {Journal of Machine Learning Research}}.

\bibitem[Griffith et~al., 2013]{Griffith2013}
Griffith, S., Subramanian, K., Scholz, J., Isbell, C.~L., and Thomaz, A.~L.
  (2013).
\newblock Policy shaping: Integrating human feedback with reinforcement
  learning.
\newblock In {\em {Neural Information Processing Systems}}.

\bibitem[Harrison et~al., 2018]{Harrison2018}
Harrison, B., Ehsan, U., and Riedl, M.~O. (2018).
\newblock Guiding reinforcement learning exploration using natural language.
\newblock In {\em Proceedings of the 17th International Conference on
  Autonomous Agents and MultiAgent Systems}, pages 1956--1958. International
  Foundation for Autonomous Agents and Multiagent Systems.

\bibitem[Hussein et~al., 2017]{Hussein2017}
Hussein, A., Gaber, M.~M., Elyan, E., and Jayne, C. (2017).
\newblock Imitation learning: A survey of learning methods.
\newblock {\em ACM Computing Surveys (CSUR)}, 50(2):21.

\bibitem[Kartoun et~al., 2010]{Kartoun2010}
Kartoun, U., Stern, H., and Edan, Y. (2010).
\newblock A human-robot collaborative reinforcement learning algorithm.
\newblock {\em Journal of Intelligent \& Robotic Systems}, 60(2):217--239.

\bibitem[Konidaris and Barto, 2007]{Konidaris2007}
Konidaris, G. and Barto, A.~G. (2007).
\newblock Building portable options: Skill transfer in reinforcement learning.
\newblock In {\em IJCAI}, volume~7, pages 895--900.

\bibitem[MacGlashan et~al., 2017]{Macglashan2017}
MacGlashan, J., Ho, M.~K., Loftin, R., Peng, B., Wang, G., Roberts, D.~L.,
  Taylor, M.~E., and Littman, M.~L. (2017).
\newblock Interactive learning from policy-dependent human feedback.
\newblock In {\em Proceedings of the 34th International Conference on Machine
  Learning-Volume 70}, pages 2285--2294. JMLR. org.

\bibitem[Madden and Howley, 2004]{Madden2004}
Madden, M.~G. and Howley, T. (2004).
\newblock Transfer of experience between reinforcement learning environments
  with progressive difficulty.
\newblock {\em Artificial Intelligence Review}, 21(3-4):375--398.

\bibitem[Parisotto et~al., 2015]{Parisotto2015}
Parisotto, E., Ba, J.~L., and Salakhutdinov, R. (2015).
\newblock Actor-mimic: Deep multitask and transfer reinforcement learning.
\newblock {\em arXiv preprint arXiv:1511.06342}.

\bibitem[Pirotta et~al., 2013]{Pirotta2013}
Pirotta, M., Restelli, M., Pecorino, A., and Calandriello, D. (2013).
\newblock Safe policy iteration.
\newblock In {\em International Conference on Machine Learning}, pages
  307--315.

\bibitem[{Plisnier} et~al., 2019]{Plisnier2019}
{Plisnier}, H., {Steckelmacher}, D., {Roijers}, D.~M., and {Now{\'e}}, A.
  (2019).
\newblock {The Actor-Advisor: Policy Gradient With Off-Policy Advice}.
\newblock {\em arXiv e-prints}, page arXiv:1902.02556.

\bibitem[Ravindran and Barto, 2003]{Ravindran2003}
Ravindran, B. and Barto, A.~G. (2003).
\newblock Relativized options: Choosing the right transformation.
\newblock In {\em Proceedings of the 20th International Conference on Machine
  Learning (ICML-03)}, pages 608--615.

\bibitem[Rusu et~al., 2015]{Rusu2015}
Rusu, A.~A., Colmenarejo, S.~G., Gulcehre, C., Desjardins, G., Kirkpatrick, J.,
  Pascanu, R., Mnih, V., Kavukcuoglu, K., and Hadsell, R. (2015).
\newblock Policy distillation.
\newblock {\em arXiv preprint arXiv:1511.06295}.

\bibitem[{Steckelmacher} et~al., 2019]{Steckelmacher2019}
{Steckelmacher}, D., {Plisnier}, H., {Roijers}, D.~M., and {Now{\'e}}, A.
  (2019).
\newblock {Sample-Efficient Model-Free Reinforcement Learning with Off-Policy
  Critics}.
\newblock {\em arXiv e-prints}, page arXiv:1903.04193.

\bibitem[Sutton et~al., 2000]{Sutton2000}
Sutton, R., McAllester, D., Singh, S., and Mansour, Y. (2000).
\newblock {Policy Gradient Methods for Reinforcement Learning with Function
  Approximation}.
\newblock {\em {Neural Information Processing Systems}}.

\bibitem[Sutton and Barto, 2018]{Sutton2018}
Sutton, R.~S. and Barto, A.~G. (2018).
\newblock {\em {Reinforcement Learning: An Introduction}}.
\newblock MIT Press, Cambridge.

\bibitem[Sutton et~al., 1999]{Sutton1999}
Sutton, R.~S., Precup, D., and Singh, S. (1999).
\newblock Between mdps and semi-mdps: A framework for temporal abstraction in
  reinforcement learning.
\newblock {\em Artif. Intell.}, 112(1-2):181--211.

\bibitem[Taylor and Stone, 2007]{Taylor2007b}
Taylor, M.~E. and Stone, P. (2007).
\newblock Cross-domain transfer for reinforcement learning.
\newblock In {\em Proceedings of the 24th international conference on Machine
  learning}, pages 879--886. ACM.

\bibitem[Taylor and Stone, 2009]{Taylor2009}
Taylor, M.~E. and Stone, P. (2009).
\newblock Transfer learning for reinforcement learning domains: A survey.
\newblock {\em {Journal of Machine Learning Research}}.

\bibitem[Taylor et~al., 2007]{Taylor2007a}
Taylor, M.~E., Stone, P., and Liu, Y. (2007).
\newblock Transfer learning via inter-task mappings for temporal difference
  learning.
\newblock {\em Journal of Machine Learning Research}, 8(Sep):2125--2167.

\bibitem[Zhan and Taylor, 2015]{Zhan2015}
Zhan, Y. and Taylor, M.~E. (2015).
\newblock Online transfer learning in reinforcement learning domains.
\newblock In {\em 2015 AAAI Fall Symposium Series}.

\end{thebibliography}

\end{document}